\documentclass[conference]{IEEEtran}

\usepackage{graphicx}
\usepackage{multirow}
\usepackage{subfig}
\usepackage{booktabs}
\usepackage{amsmath}
\usepackage{float}

\ifCLASSOPTIONcompsoc
  \usepackage[nocompress]{cite}
\else
  \usepackage{cite}
\fi

\ifCLASSINFOpdf

\else

\fi

\begin{document}
%
\title{Identifying Diabetic Patients with High Risk of Readmission}

\author{
    \IEEEauthorblockN{Bhuvan M S\IEEEauthorrefmark{1}, Ankit Kumar\IEEEauthorrefmark{2}, Adil Zafar\IEEEauthorrefmark{3}, Vinith Kishore\IEEEauthorrefmark{1}}
    \IEEEauthorblockA{\IEEEauthorrefmark{1}National Institute of Technology Karnataka, Surathkal, India
    \{msbhuvanbhuvi, vinithkishore\}@gmail.com}
    \IEEEauthorblockA{\IEEEauthorrefmark{2}Indian Institute of Technology, Guwahati
    \{ankitkr710\}@gmail.com}
    \IEEEauthorblockA{\IEEEauthorrefmark{3}Indian Institute of Technology, Kharagpur
    \{adil.dba.zafar\}@gmail.com}
}

\maketitle

\begin{abstract}
Hospital readmissions are expensive and reflect the inadequacies in healthcare system. In the United States alone, treatment of readmitted diabetic patients exceeds 250 million dollars per year. Early identification of patients facing a high risk of readmission can enable healthcare providers to to conduct additional investigations and possibly prevent future readmissions. This not only improves the quality of care but also reduces the medical expenses on readmission. Machine learning methods have been leveraged on public health data to build a system for identifying diabetic patients facing a high risk of future readmission. Number of inpatient visits, discharge disposition and admission type were identified as strong predictors of readmission. Further, it was found that the number of laboratory tests and discharge disposition together predict whether the patient will be readmitted shortly after being discharged from the hospital (i.e. \textless30 days) or after a longer period of time (i.e. \textgreater30 days). These insights can help healthcare providers to improve inpatient diabetic care. Finally, the cost analysis suggests that \$252.76 million can be saved across 98,053 diabetic patient encounters by incorporating the proposed cost sensitive analysis model.
\end{abstract}


\IEEEpeerreviewmaketitle

\section{Introduction}
A survey conducted by the Agency for Healthcare Research and Quality (AHRQ) found that in the year 2011 more than 3.3 million patients were readmitted in the United States within 30 days of being discharged \cite{1}.  The need for readmission indicates that inadequate care was provided to the patient at the time of first admission. Inadequate care poses threat to patients life and treatment of readmitted patients leads to increased healthcare costs. Over 41 billion dollars were spent on treatment on readmitted patients in 2011 \cite{1}.  Diabetes is the seventh leading cause of death and affects about 23.6 million people in the United States. Hospital readmission being a major concern in diabetes care, over 250 million dollars was spent on treatment of readmitted diabetic patients in 2011 \cite{1}.  

Patients facing a high risk of readmission need to be identified at the time of being discharged from the hospital, to facilitate improved treatment to reduce the chances of their readmission. 
Readmission of patients within 30 days of being discharged (short-term readmission) has been a widely used metric for studying readmissions \cite{1}. However, a significant number of diabetic patients are readmitted after 30 days of being discharged (long-term readmission). As opposed to previous work done in the domain we consider both short-term and long-term readmission scenarios.
In addition to an effective prediciton model, identifying risk factors (features in the medical record) correlating to readmission will help in considering these factors with greater care and better documentation in future medical records, thereby developing more efficient medical protocols.

The core idea is to provide a comprehensive data solution to readmission problem to facilitate implementation at the healthcare institutions to embark a significant improvement in the in-patient diabetic care. This solution provides all round information to the implementing healthcare institution along with the cost analysis model. 

Addressing these critical problems involves several data challenges which are considered throughout the research. The main contribution of this work are:
\begin{itemize}
\item Prediction of diabetis patients with high risk of readmission, by modeling multivariate patient medical records using machine learning classifiers. Incorporating a conservative prediction model to have higher recall as suited to healthcare institutions.
\item Analysis of characteristics of Short-term (within 30 days) and Long-term (after 30 days) readmissions with differeent classifiers.
\item Identifying the critical risk factors (features) using ablation study.
\item Identifying the association across critical risk factors using association rule mining.
\item Cost analysis to determine the effective cost saved by implementing the work in real world.
\end{itemize}

The rest of the paper is organized as follows: Section 2 presents a brief overview of the past work. Section 3 describes the dataset used and the proposed methodology covering all the enumerated points mentioned above. Results and discussions are presented in Section 4 with respected to each part, followed by conclusion and future work in section 5.

\section{Related Work}
Numerous previous studies have analyzed the risk factors that predict readmissions rates of diabetic patients \cite{1, 2, 3, 19, 20, 21, 22, 23, 24, 25, 26}, out of which significant ones have been discussed here.  \cite{19} Found that acute and chronic glycemic control influenced readmission risk in a dataset of more than 29,000 patients over the age of 65. \cite{21} Analyzed the readmission risk for a dataset of more than 52,000 patients in the Humedica network. \cite{26} Studied demographic and socioeconomic factors which influence readmission rates. \cite{2} Studied the impact of HbA1c on readmissions. \cite {generalreadmission} have conducted analysis on predictability of hospital readmissions in general without targetting any specific disease. The dataset considered in this case, covers demographic, clinical procedure-related and diagnostic-related features along with drug information for patients above 65 years of age to predict the readmission within 30 days (short-term readmission prediction). It contains comprehensive results on feature reduction methods, but the performance of the prediction models are modest. Cost analysis and grouped features importance mining was not considered and the analysis was not targeted towards specific disease like diabetis.

To the best of our knowledge, our work is the first one analyzing diabetic patients faced with the risk of both short and long term of readmission along with feature analysis and cost analysis. We use a bigger and a more balanced dataset (i.e. data across all age groups and across 130 hospitals) as compared to previous works. Consequently, our results are more reflective of the problem of readmissions among diabetic patients. 

Previous works have not documented the performance of different machine learning classifiers. Moreover, both short-term and long-term readmission scenarios were not considered. Though they solved a specific problem, they do not provide a single comprehensive solution to readmission problem that can readily be implemented. 

In addition to addressing the above gaps in the research, this work covers methods to identify the critical risk factors in predicting the readmission rates. Knowledge of such factors is likely to be useful in developing protocols for better inpatient diabetes care. We hope that results presented in this work will serve as a good baseline for any future work to compare against.

\section{Methodology}
A dataset containing medical records of 101765 diagnosed with diabetes, collected over a period of 10 years (1999-2008) from 130 hospitals in USA was used for all analysis presented in this work \cite{2, 3}.  The medical record of each patient included 50 potential risk factors and a label indicating whether the patient was readmitted within 30 days, after 30 days or was never readmitted. The distribution is as follows - 11\% of patients were readmitted within 30 days, 35\% after 30 days and 54\% patients were never readmitted. 

\begin{figure}
\centering
\includegraphics[height=3in, width=3in]{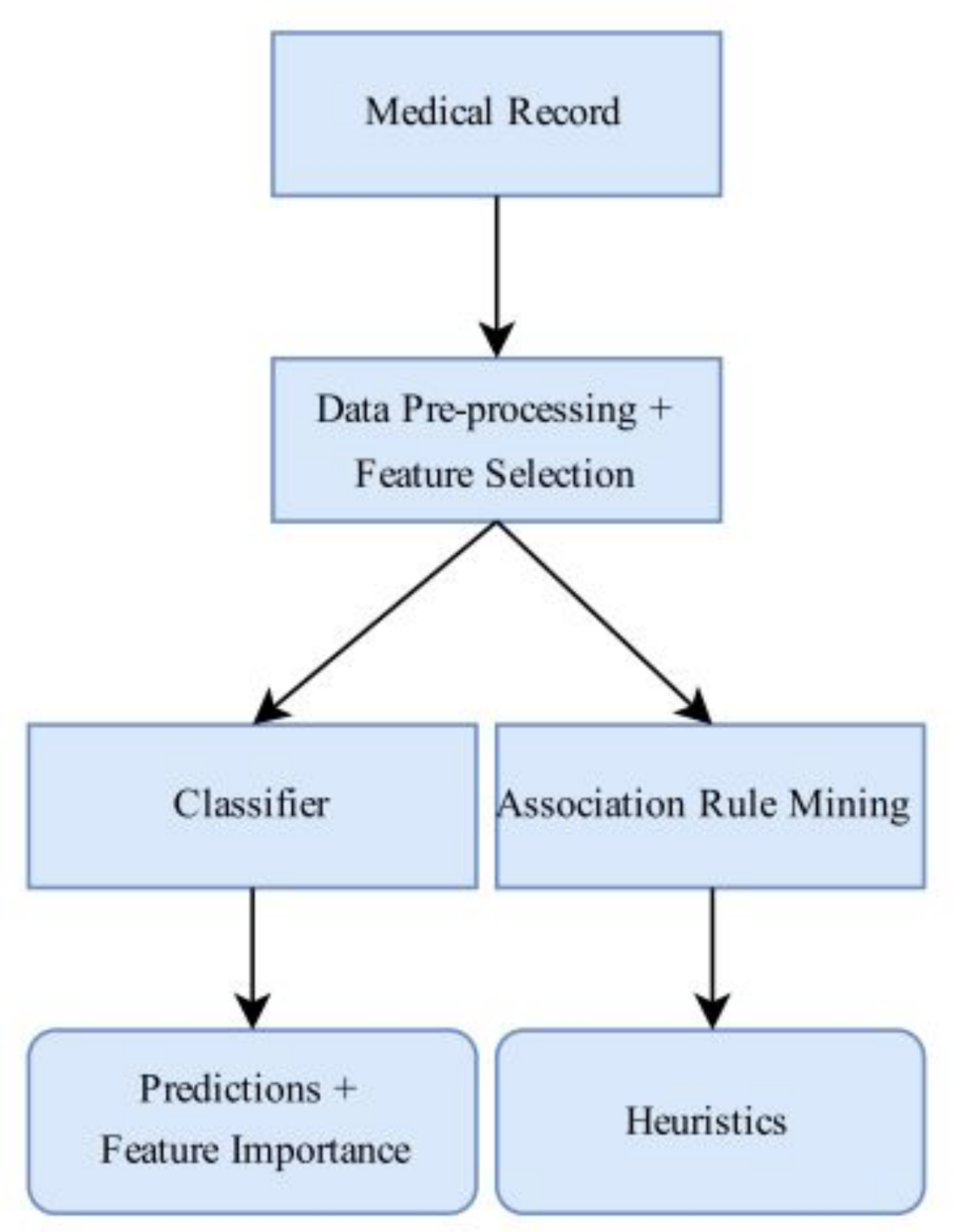}
\caption{Overview of the methodology. The data was preprocessed according to the method described in section 3.A. Using the preprocessed data we built models for identifying high-risk patients (classification, section 3.B) and identifying groups of features that were important for predicting readmission rates (feature analysis, section 3.C).}
\label{fig:flowchart}
\end{figure}

An overview of our method is provided in Figure ~\ref{fig:flowchart}.  We first preprocess the data according to the method described in section 3.A. Using this preprocessed data we build models for predicting readmission rates (classification, section 3.B) and identifying groups of features that are important for predicting readmission rates (feature analysis, section 3.C) which in turn uncover the critical risk factors.
\begin{table}
\centering
\caption{The list of risk factors considered for predicting readmission rates \cite{2}}
\begin{tabular}{|l|l|l|l|}
\hline
No. & Feature                                                                     & No. & Feature                                                                       \\ \hline
1   & Race                                                                        & 12  & \begin{tabular}[c]{@{}l@{}}Number of \\ Emergency \\ Visits (NE)\end{tabular} \\ \hline
2   & Gender                                                                      & 13  & \begin{tabular}[c]{@{}l@{}}Number of \\ Inpatient \\ Visits (NI)\end{tabular} \\ \hline
3   & Age                                                                         & 14  & \begin{tabular}[c]{@{}l@{}}Diagnosis 1 \\ (Primary) (PD)\end{tabular}         \\ \hline
4   & \begin{tabular}[c]{@{}l@{}}Admission \\ Type (AT)\end{tabular}              & 15  & \begin{tabular}[c]{@{}l@{}}Diagnosis 2 \\ (Secondary) (SD)\end{tabular}       \\ \hline
5   & \begin{tabular}[c]{@{}l@{}}Discharge \\ Disposition (DD)\end{tabular}       & 16  & \begin{tabular}[c]{@{}l@{}}Diagnosis 3 \\ (Tertiary) (TD)\end{tabular}        \\ \hline
6   & \begin{tabular}[c]{@{}l@{}}Admission \\ Source (AS)\end{tabular}            & 17  & \begin{tabular}[c]{@{}l@{}}Number of \\ Diagnoses (ND)\end{tabular}           \\ \hline
7   & \begin{tabular}[c]{@{}l@{}}Time in \\ Hospital (Days)\end{tabular}          & 18  & \begin{tabular}[c]{@{}l@{}}Glucose \\ Serum Test (GST)\end{tabular}           \\ \hline
8   & \begin{tabular}[c]{@{}l@{}}Number of \\ Lab Procedures\end{tabular}         & 19  & A1C Test Result                                                               \\ \hline
9   & \begin{tabular}[c]{@{}l@{}}Number of \\ Procedures\end{tabular}             & 20  & Insulin                                                                       \\ \hline
10  & \begin{tabular}[c]{@{}l@{}}Number of \\ Medications\end{tabular}            & 21  & \begin{tabular}[c]{@{}l@{}}Change of \\ Medication\end{tabular}               \\ \hline
11  & \begin{tabular}[c]{@{}l@{}}Number of \\ Outpatient Visits (NO)\end{tabular} & 22  & \begin{tabular}[c]{@{}l@{}}Diabetic \\ Medication (DM)\end{tabular}           \\ \hline
\end{tabular}
\label{tab:features}
\end{table}

\subsection{ Risk Factors}
Prior to performing any analysis we processed the data in the following way: The primary, secondary and tertiary medical diagnoses were indicated by the ICD9 codes \cite{7}. Each ICD9 indicates a unique diagnostic condition. The ICD9 codes took more than 1000 unique values. For many diagnostic conditions (ICD9 codes) data were available for only a few patients. Thus, determining the effect of each diagnostic condition on the readmission rates was not feasible. Consequently, we grouped ICD9 codes representing similar diagnoses into a total of 10 groups \cite{7}. If grouping was not done, each of these ICD-9 codes would be present in smaller number of samples. Hence, individually each of these ICD-9 codes would not significantly represent the data and might receive a lower importance weight. Moreover, many of these codes are very closely related hence they can be grouped meaningfully as explained in \cite{7}. Since each ICD-9 code indicates a single medical complication, grouping all related complications (for e.g. grouping all complications related to respiratory system from codes 450-519) into one nominal feature value is legitimate and meaningful to consider in a healthcare scenario. This would be crucial to determine the effect of the diagnostic condition on the readmission rate. The dataset also provided details of the medication administered to each patient. We found that the only medication that varied across the patients was the delivery of insulin, while other medications remained common among all the patients hence only insulin medication was considered as a feature. For some factors such as weight, payer code and medical specialty – the data were missing for 97\%, 40\% and 49\% of the patients respectively. Consequently, we ignored these factors in our analysis. Future surveys and data collections need to collect these information to help the analysis. The race of the patient and the type of diagnoses were missing for 2\% and 1\% of the patients considered in the study. We removed such patients from the dataset.  After this data pre-processing we were left with 22 factors listed in Table ~\ref{tab:features}. The detailed definitions of these factors can be found in \cite{2}.

Some factors such as the type of diagnoses and admission type took nominal values where as others such as number of inpatient visits and time in hospital took numerical values. While some algorithms for predicting risk of readmission rates naturally deal with nominal and numerical data both (e.g. random forests), other algorithms such as neural networks cannot deal with nominal values.  For algorithms that only operate on numerical data, nominal values were converted into binary features. For example, the factor 'Admission Type' that took one out nine distinct values was represented by a 9-D binary vector. Each dimension was set to 1 or 0 depending on the value that Admission type took.

\subsection{ Classification}
Identification of high-risk diabetic patients was posed as the problem of classifying of whether a patient would be readmitted within 30 days of being discharge or after 30 days of being discharged or never readmitted. Since, different classification algorithms are apt for different kinds of data we experimented and compared results from five different algorithms. Prior to training the classification algorithms, we randomly split our dataset into two distinct sets - the training and the test set. The training and test set consisted of 75\% and 25\% of the data.  The parameters of each algorithm were chosen based on the classification performance evaluated by five-fold-cross-validation on the training set. The performance of all algorithms was evaluated on the test set. 

A short description of all classification algorithms considered in this work is provided below.
\subsubsection{	Naive Bayes}
Naive Bayes algorithm \cite{8} is a probabilistic model for classification. It assumes that given the class, features are statistically independent of each other. 

\subsubsection{ Bayesian Networks}
Bayesian networks \cite{10} estimate the probability distribution of a class by modeling the relationship between features by a acyclic undirected graph (in general Bayes networks can be directed and cyclic, but for our experiments we only considered acyclic and undirected models). 

\subsubsection{ Random Forest}
Random Forest is composed of a set of decision trees. Each decision tree acts as a weak classifier and pooling the responses from multiple decision trees leads to a strong classifier (random forest \cite{12}). Each decision tree is trained independently and determines the class of an input by evaluating a series of greedily learned binary questions. The random forest consisting of 250 trees, each of depth of utmost 5 nodes was used, as it was found to be optimal from the experiment with varying number of trees and depth in the forest. 

\subsubsection{ Adaboost}
Adaboost constructs a strong classifier by sequentially combining a set of weak classifiers \cite{13}. At first iteration, a single classifier is learnt to minimize the classification error. At each consequent iteration, a new classifier is learnt which seeks to minimize the error of the classifier composed of the set of classifiers learnt until the previous iteration. In all our experiments, decision trees were used as weak classifiers.

\subsubsection{ Neural Networks}
Neural Networks \cite{28} are one of the powerful classifiers which has established state of art results in like speech processing, computer vision \cite{29} and a wide variety of other tasks. We have used a MultiLayer Perceptron (MLP) which is a feed-forward artificial neural network \cite{mlpc} model that maps sets of input data onto a set of appropriate outputs. A MLP consists of multiple layers of nodes in a directed graph, with each layer fully connected to the next one. Except for the input nodes, each node is a neuron (or processing element) with a non-linear activation function. MLP is a modification of the standard linear perceptron and can distinguish data that are not linearly separable. It is trained with one hidden layer by minimizing the squared error plus a quadratic penalty with the BFGS method \cite{bfgs}. We experimented with Neural Networks with one hidden layer consisting of 2, 4 and 8 nodes and observed that there was very small increase in the performance but was not significant to consider. Hence considering the training time of these neural networks we decided to use Neural Network with one hidden layer of 2 nodes for further experiments. We used Weka 3.7 \cite{9} for training neural networks using MLP Classifier \cite{15}.

\subsection{ Feature Analysis}
Various features in the medical records are analyzed to get insights about their importance in predicting the readmission status of a patient encounter instance. Following subsections explain two methods of carrying the feature analysis to discover critical risk factors.

\subsubsection{ Ablation Study of Risk Factors}
The importance of individual risk factors can be judged by performing an ablation study. An ablation study involves removing one factor at a time and comparing the accuracy of predicting readmission with this set of features with the accuracy obtained by considering all the features. Intuitively, removal of more important features should lead to larger decreases in accuracy. Ablation of risk factors was performed to judge the importance of each of these factors.

\subsubsection{ Associative Rule Mining (ARM)}
Along with the predictive modeling the main theme is to identify the risk factors and help medical practitioners get more insights. One of the objectives is to assist the practitioners and researchers in getting more insights to understand why certain patients get readmitted and factors leading to it, hence indirectly increasing the efficiency of the diagnosis in order to prevent further readmissions. Ablation study helps to understand the risk factors, which is further supported by significant association rules.

Groups of consistently occurring factors that influence readmission rates can be revealed by associative rule mining (ARM; Apriori algorithm; \cite{18}).  ARM aims at identifying rules of the form A =\textgreater B where A is a conjunction of an arbitrary number of factors and B is a factor that is predicted to occur if A is true. The tuple (A, B) forms a set of factors which commonly appear in the dataset. Discovering the groups of factors that commonly occur among readmitted patients or patients those are never readmitted can further our understanding about causes of readmission. If for a patient factors A and B both were present, we deemed that the patient followed the rule (A, B). We determined such rules by mining the medical records of all patients in our dataset. Next, we performed class sensitive ARM – where we took examples either from readmitted patients or patients that were never readmitted.  Then for each rule, we determined the number of patients in the entire dataset that followed this rule. Within this set, we computed the fraction of patients which were readmitted \textless30 days and the fraction that were readmitted at any time in the future. All the rules were then sorted based on the fraction of patients which followed the rule and were readmitted within 30 days. The rules with the highest fraction of readmissions indicate the factors that are strong predictors of risk of readmission. The rules with lowest fraction of readmissions indicate the factors that are strong predictors of low risk of readmission.

\subsection{ Evaluation Criteria}
A system for predicting high-risk patients is only useful, if a large fraction of patients at high-risk are correctly identified (i.e. high recall) without raising a large number of false alarms (i.e. high precision). Receiver Operating Characteristics (ROC) is ubiquitously known as one of the best metrics to evaluate classification models. The research paper  \cite{roc_pr_relationship}, clearly proves that a deep connection exists between ROC space and Precision-Recall (PR) space, such that a curve dominates in ROC space if and only if it dominates in PR space. The paper also states that, though ROC curves are commonly used to present results for binary decision problems in machine learning, when dealing with highly skewed datasets (like our dataset), PR curves give a more informative picture of an algorithms performance. Therefore, we chose to present PR curves. Moreover PR curves provide sufficient information to select the appropriate threshold needed to tune for higher recall which is necessary in a healthcare scenario (a conservative scenario). All methods presented in this paper were evaluated based on recall and precision. The definition of these evaluation metrics is provided below:
\begin{itemize}
\item Precision (P): is defined as the fraction of ground truth positives among all the examples that are predicted to be positive.
\begin{equation}
Precision = TP/ (TP + FP)
\end{equation}
Where, TP, FP stand for True Positives and False Positives respectively
\item Recall (R): is defined as the fraction of ground truth positives that are predicted to be positives.
\begin{equation}
Recall = TP/ (TP + FN)
\end{equation}
Where, FN denotes False Negatives.
\item Precision-Recall Curve \cite{27}:  The tradeoff between the precision and recall can be studied by looking at the plot of how precision changes as a function of recall. This plot is known as the Precision-Recall (PR) curve. The precision is plotted on the y-axis and recall on x-axis. The area under precision recall curve is a cost-effective metric for evaluation of classifiers. High value of area under the PR curve indicates better performance. Area under the PR curve is preferred way of studying the performance of classifiers in skewed (i.e. imbalance between positives and negatives) datasets such as the one considered in this work \cite{27}. 
\end{itemize}

\subsection{ Cost Sensitive Analysis}
The model needs to be tuned to maximize the cost saved by the hospital by using the analysis model. This can be done by selecting appropriate threshold for the machine learning algorithms. 
Let \(\alpha\) be the cost incurred per readmission, and \(\beta\) be the cost per Special Diagnosis for Patients predicted as Yes (\textless30 or \textgreater30 day Readmissions). Let the patient encounter instances of the test set be defined according to the below cost matrix. Cost matrix is one in which cost or penalty of classification will be specified for each element as in the confusion matrix.

\begin{equation}
Cost Matrix = 
\begin{bmatrix}
TP    &     FN    \\
FP    &     TN    \\
\end{bmatrix}
\end{equation}

Where TP, FP, FN and TN correspond to True Positives, False Positives, False Negatives and True Negatives respectively.

Without the prediction model: all the instances where patients actually get readmitted (True Positives and False Negatives) incur a cost of \(\alpha\), hence defined by the below matrix.
cost matrix without prediction model.

\begin{equation}
Without Prediction Model (C) = \begin{bmatrix}
\alpha      &     \alpha      \\
0     &     0     \\
\end{bmatrix}
\end{equation}

With the prediction model: considering that all patients who are predicted to get readmitted (True Positives and False Positives) would be examined with a special diagnosis which would cost \(\beta\), which in turn would prevent their readmission. Hence by above consideration, we make a realistic hypothesis that the special diagnosis provided for predicted instances prevent their readmission (Here the prediction is helping to determine whom to provide special diagnosis). The patient encounters who are predicted to be readmitted but who do not actually get readmitted, would also incur a cost of beta for special diagnosis, which will just serve as preventive measure in conservative hospital scenario. Cost of \(\alpha\) would be incurred for those patients who are predicted not to get readmitted but who actually get readmitted. Those patients who where predicted as not readmitted and those who actually don't get readmitted (True Negatives) anyway would not contribute to any cost, but it may simply help the hospital for capacity planning. Hence with the predictive model we would get the cost matrix C' shown below. In our research special diagnosis is considered to be 'One Extra Day' during the initial admission during which the doctor can conduct another diagnosis which might result in the discovery of new complications in the patient. Hence the cost of special diagnosis is considered to be cost per one day of admission in the hospital.

\begin{equation}
With Prediction Model (C') = \begin{bmatrix}
\beta &     \alpha      \\
\beta &     0     \\
\end{bmatrix}
\end{equation}

The difference of cost matrix without and with prediction model gives the saved cost matrix as shown below.
\begin{equation}
Saved Cost Matrix (S) = C - C'
= \begin{bmatrix}
\alpha - \beta    &     0     \\
-\beta      &     0     \\
\end{bmatrix}
\end{equation}

The prediction model needs to be tuned to appropriate thresholds to maximize this saved-cost matrix, hence maximizing the cost saved by the analysis model.

\section{Results and Discussions}
This section contains results consolidated from carious experiments as described in the section 3. The effectiveness of various classifiers on predicting the readmission status have been discussed in the section 4.A. The section 4.B describes critical risk factors that have been identified by feature analysis conducted by two methods namely, the ablation study and Association Rule Mining. Finally, the results of cost effective analysis have been discussed in section 4.C.

\subsection{ Analysis of Classifiers}
\begin{figure}
\centering
\includegraphics[height=2.65in, width=3.5in]{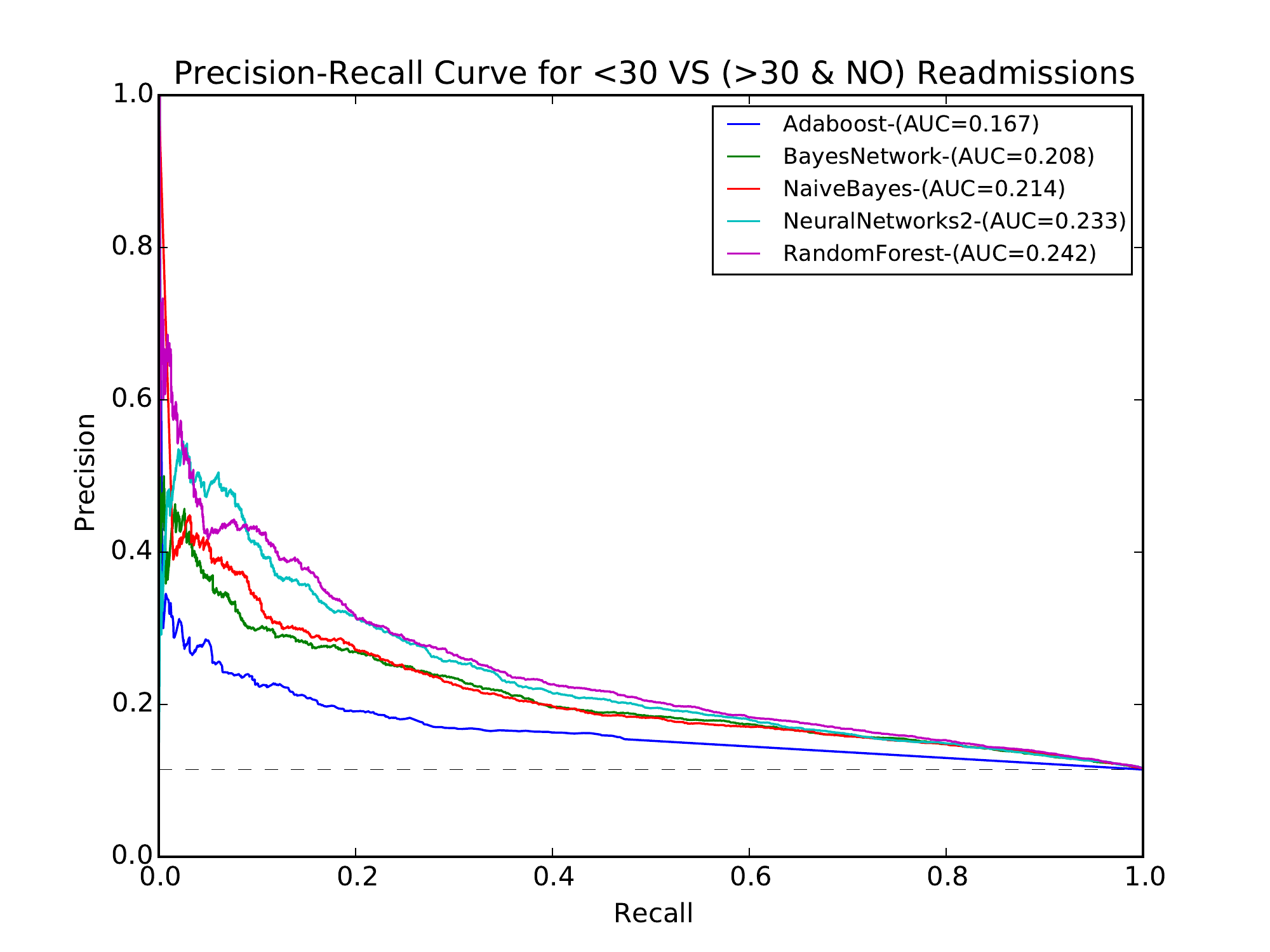}
\caption{Comparing the accuracy of different methods for identifying high-risk patients that are readmitted within 30 days of being discharged. The dotted line shows the performance of a classifier at chance accuracy.}
\label{fig:prl30}
\end{figure}

\begin{figure}
\centering
\includegraphics[height=2.65in, width=3.5in]{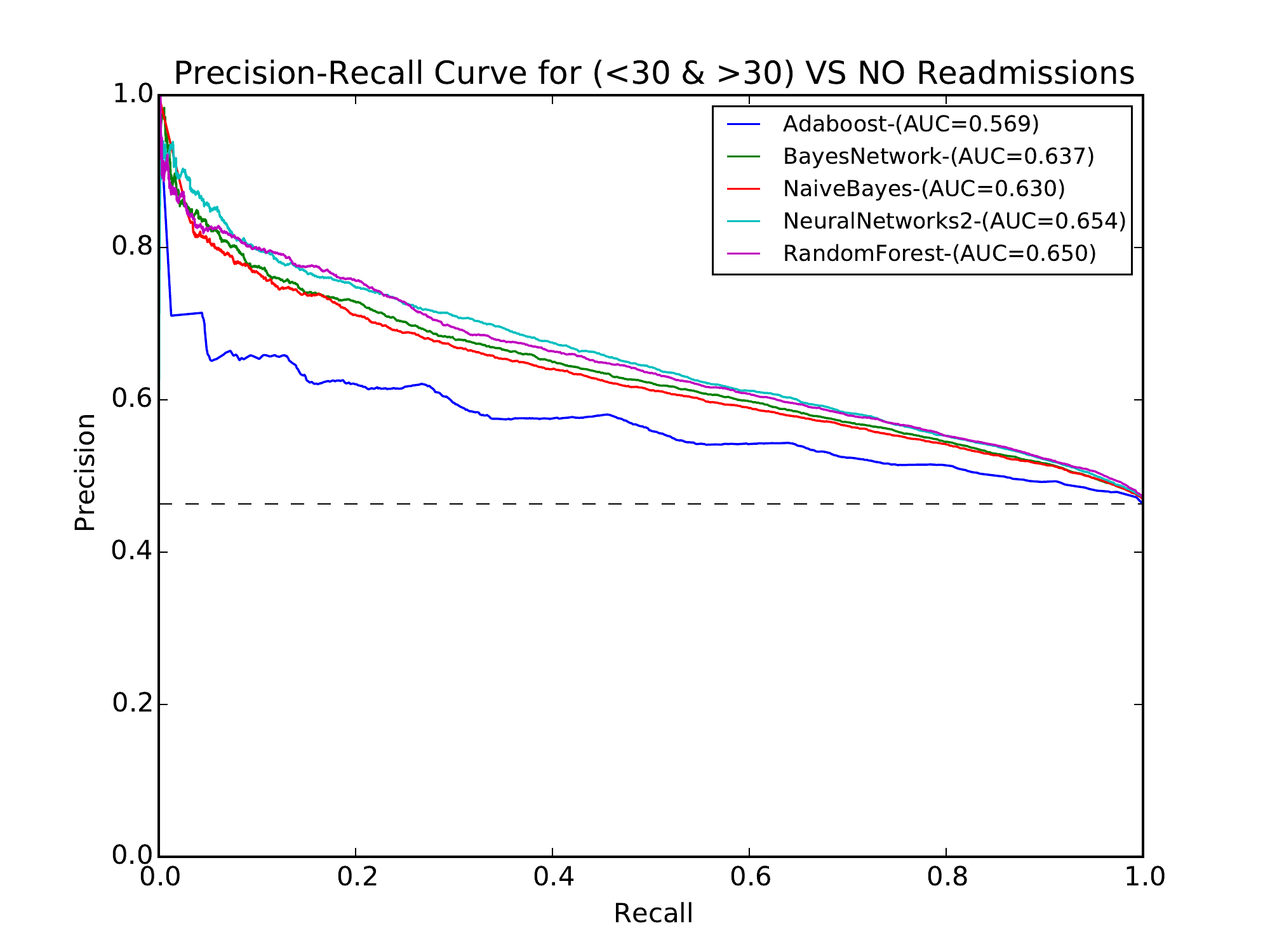}
\caption{Comparing the accuracy of different methods for identifying high-risk patients that are readmitted in the future. The dotted line shows the performance of a classifier at chance accuracy.}
\label{fig:prl30g30}
\end{figure}

The distribution of patient encounters is skewed because; most patients in the dataset were never readmitted (54\%). Only 11\% of patients were readmitted within 30 days(\textless30), while the rest (35\%) of them were readmitted after 30 days. But the greater than 30 day readmission class (\textgreater30) is ambiguous as it could mean 31 days or even a couple of years. In first case, by making an hypothesis that '\textgreater30 day readmission' instances might have patterns similar to 'NO readmission' instances, we combined \textgreater30 and NO readmission instances and built a binary classification model to classify \textless30 versus (\textgreater30 \& NO),i.e., identifying high risk patients readmitted within 30 days of being discharged (Figure ~\ref{fig:prl30}). In the second case, we made a hypothesis readmission either after \textless30 and \textgreater30 days might have similar patterns and combined \textless30 days and \textgreater30 days readmission instances and built a binary classification model to classify Readmission (\textless30 \& \textgreater30) versus 'NO readmission', i.e., identifying high risk patients readmitted at any time in the future (Figure ~\ref{fig:prl30g30}). In this case, we can see that distribution is almost evened for the two classes. 

We experimented with both the cases. As we observe from Table ~\ref{tab:classifiercomp}, area under PR curves for best performing algorithm in each case came out to be higher in the latter case i.e., while identifying high risk patients readmitted at any time in the future. Therefore, the performance of identifying high-risk readmitted patients within 30 days is lower than identifying high-risk patients readmitted at any time in the future.

\begin{table}
\centering
\caption{Comparing the performance of different algorithms for identifying high risk patients}
\begin{tabular}{|l|l|l|}
\hline
\multirow{2}{*}{Classifier}                                 & \multicolumn{2}{l|}{Area Under Precision-Recall Curve}       \\ \cline{2-3} 
                                                            & Class \textless30 & Class \textless30 + Class \textgreater30 \\ \hline
\begin{tabular}[c]{@{}l@{}}Naive Bayes\end{tabular}     & 0.214             & 0.63                                     \\ \hline
\begin{tabular}[c]{@{}l@{}}Bayes Network\end{tabular}   & 0.208             & 0.637                                    \\ \hline
\begin{tabular}[c]{@{}l@{}}Random Forest\end{tabular}   & 0.242             & 0.65                                     \\ \hline
\begin{tabular}[c]{@{}l@{}}Adaboost Trees\end{tabular}  & 0.167             & 0.569                                    \\ \hline
\begin{tabular}[c]{@{}l@{}}Neural Networks\end{tabular} & 0.233             & 0.654                                    \\ \hline
\end{tabular}
\label{tab:classifiercomp}
\end{table}

\subsection{ Identifying the critical risk factors}
Feature analysis described in the section 3.C leads to the discovery of importance of risk factors estimated by two methods presented in the following subsections.

\subsubsection{ Using Ablation Study}

\begin{figure*}
\centering
\includegraphics[height=3in, width=7in]{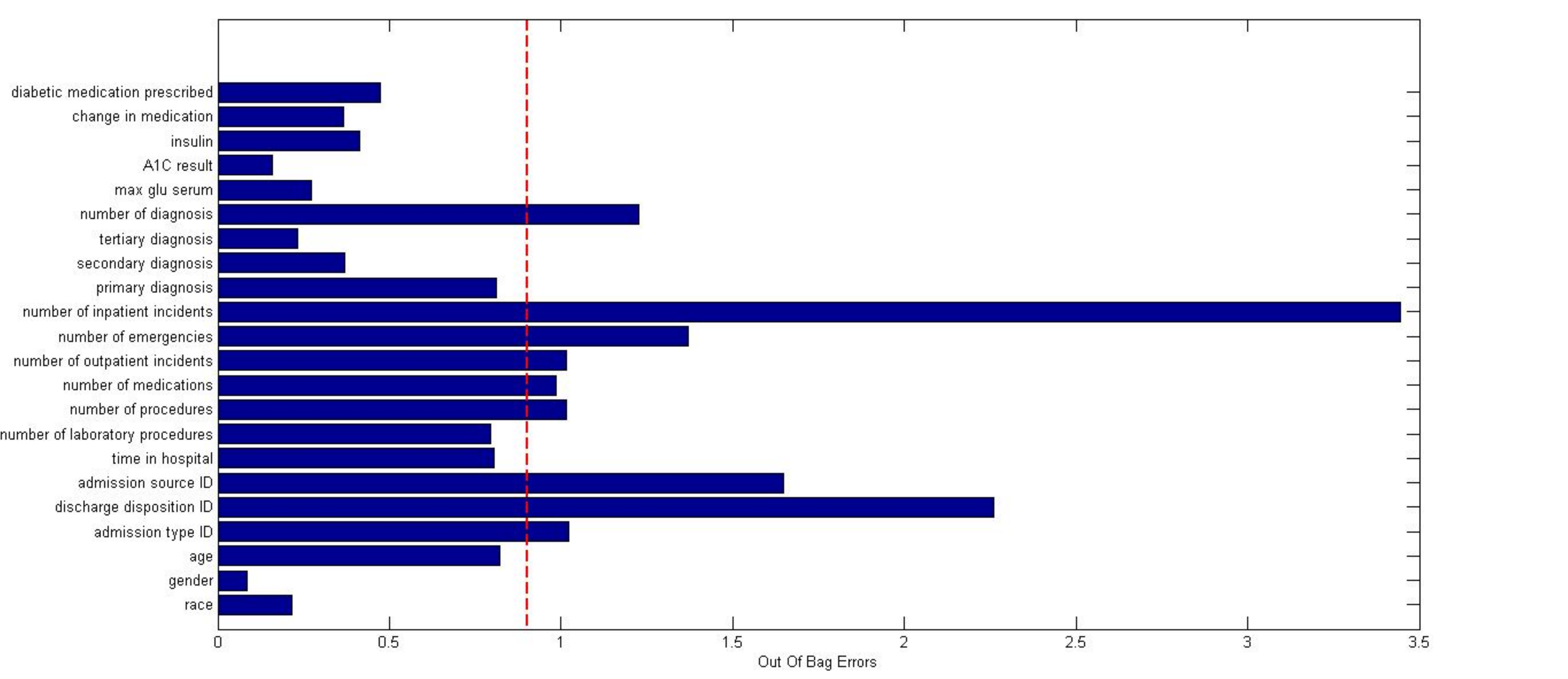}
\caption{Analyzing the importance of individual risk factors for identifying high-risk patients. The importance of each risk factor was estimated by computing the out of bag error (Section 3.C.1). Larger out of bag error indicates that the risk factor is more important. The number of inpatient incidents, the discharge disposition and admission type are most important for identifying high risk patients.}
\label{fig:featureimphighrisk}
\end{figure*}

\begin{figure*}
\centering
\includegraphics[height=3in, width=7in]{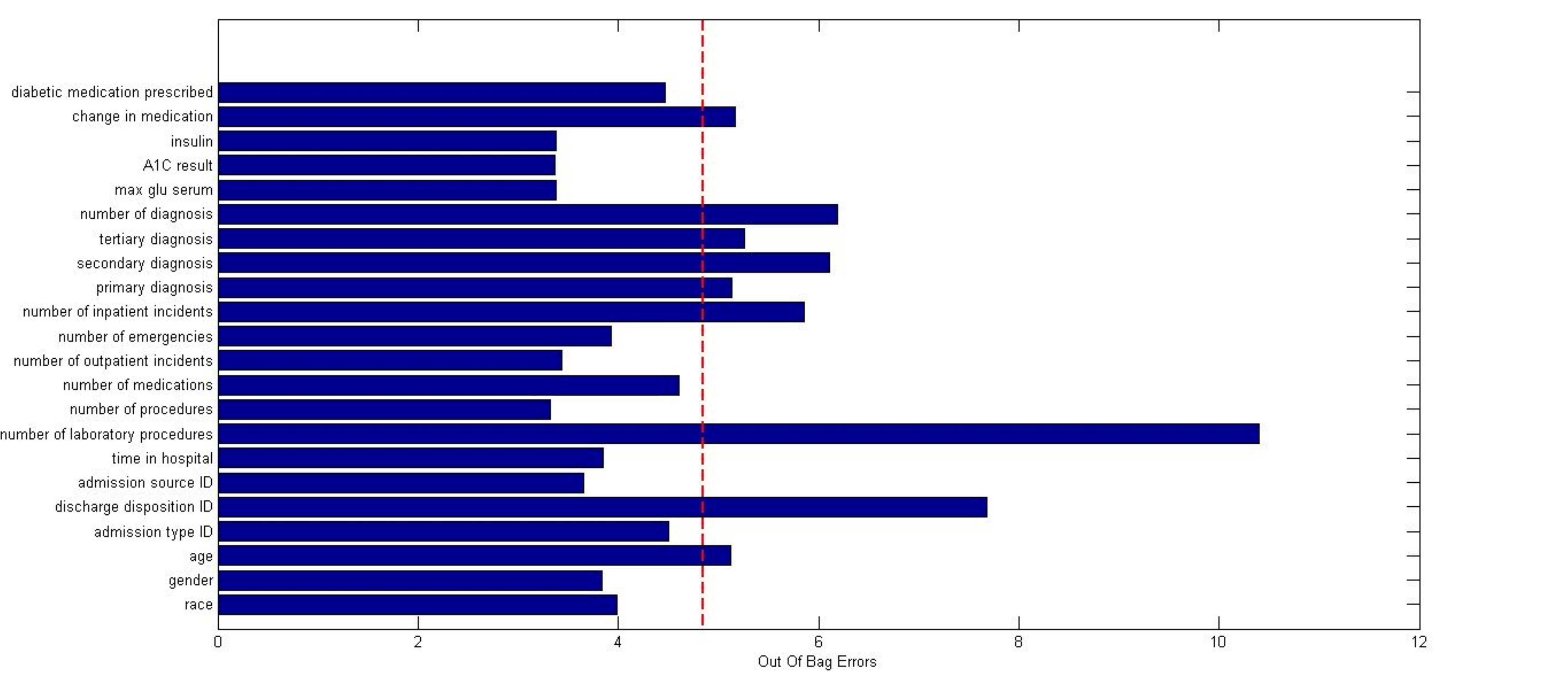}
\caption{Analyzing the importance of individual risk factors in differentiating risk factors influencing short-term and long-term readmissions. The importance of each risk factor was estimated by computing the out of bag error (Section 3.C.1). Larger out of bag error indicates that the risk factor is more important. The number of laboratory procedures and discharge disposition are found to be most important for differentiating short-term readmissions from long-term readmissions.}
\label{fig:featureimpdifferentiatingrisk}
\end{figure*}

Random Forests were most accurate at identifying patients with high-risk of readmissions. This suggested that random forests could be used to identify the importance of each risk factor in identifying high-risk patients. Such factors were identified using an ablation study (section 3.C.1). The ablation study compares the performance of classifiers first by considering all the features, and then removing one of these features. The difference in the accuracies is used as the estimate of importance of the feature. Figure ~\ref{fig:featureimphighrisk} shows the increase in out of bag errors (i.e. increase in inaccuracy), for each factor. We observed that the number of inpatient visits, discharge disposition and admission type are most important for identifying high-risk patients.

Further, in order to gain additional insights into causes of readmissions, we sought to analyze if there were any differences between the factors that led to readmission within 30 days and after 30 days. For this we trained a random forest for differentiating patients readmitted within 30 days from the patients who were readmitted after 30 days. We then performed an ablation study on identify the factors which were different among short-term and long-term readmissions. The results of our analysis are presented in Figure ~\ref{fig:featureimpdifferentiatingrisk}. We found that the number of lab tests (i.e. laboratory procedures) is useful in differentiating between short and long term readmissions. We also found that patients who are discharged to home are more likely to be readmitted within 30 days as opposed to being readmitted after 30 days.

\subsubsection{ Association Rule Interpretation}
Apriori algorithm retrieved several general and class association rules. These association rules by themselves are a treasure which would help doctors for improving efficiency of initial diagnosis. Prominent Association Rules have been given in Table ~\ref{tab:par} with respective support and confidence. More of these rules have been put up in appendix A. These rules can be interpreted by inference logic from the Consequent (LHS) to Antecedent (RHS).

The rule 3 in Table ~\ref{tab:par}, indicates that Caucasian female patients who were primarily diagnosed with Caucasian female patients who were primarily diagnosed with Dyspnea and Respiratory abnormalities (ICD9-786.0), and had Hypertension and Maligancy (ICD9-401.0) as secondary diagnosis, are less probable (2.27\%) to be readmitted within 30 days, as only 5 such patients out of 220 were readmitted within 30 days during the survey. Similarly rule 6 indicates that, patients admitted through emergency (admission source id=7), and discharged or transferred to Skilled Nursing Facility (discharge disposition id=3) are slightly more probable (\begin{math}15.19 + 36.97 = 52.16\%\end{math}) to get readmitted within or after a month. Similarly, the rule 1 implies that a patient encounter who is diagnosed with Cellulitis/abscess, face (ICD-9: 682.0) and Diabetes mellitus without mention of special complication (ICD-9: 250.0) as primary and secondary diagnosis respectively has low possibility of getting readmitted within 30 days (1.88\%) and mostly would not get readmitted at any time in future (71.83\%).  

When the prediction model predicts that this particular patient will not get readmitted, as a medical practitioner, one can use this matched rule to statistically understand that similar patients (who had similar diagnosis) were not readmitted in the past. It might help him to get him more insights and would assist him in further diagnosis. He would understand that such patients are less prone to risk. Such interpretations from all the rules could be very useful
for taking informed decisions and gauging the risk factor. We observe that most of these rules can only help to decide about the probability of patients not getting readmitted within 30 days, due to the biased distribution of dataset.

\begin{table*}[h]
\centering
\caption{Prominent Association Rules (Refer Tables ~\ref{tab:appendixicd9} and ~\ref{tab:appendixmapping} in the appendix for ICD-9 code and Id Mappings respectively)}
\begin{tabular}{|l|l|l|l|l|l|}
\hline
\multirow{3}{*}{No.} & \begin{tabular}[c]{@{}l@{}}Readmission\\   Status\end{tabular}               & \textless30     & \textgreater30     & NO       & TOTAL         \\ \cline{2-6} 
                     & Total Number of Instances                                                    & 11357           & 35545              & 54863    & 101765        \\ \cline{2-6} 
                     & Association Rules (Consequents)                                              & \multicolumn{3}{l|}{Class-Wise Matches (in \%)} & Total Matches \\ \hline
1                    & diag\_1=682.0; diag\_2=250.0                                                 & 1.88            & 26.29              & 71.83    & 213           \\ \hline
2                    & diag\_1=786.0; diag\_2=250.0; diag\_3=401.0                                  & 2.16            & 22.51              & 75.32    & 231           \\ \hline
3                    & race=Caucasian; gender=Female; diag\_1=786.0; diag\_3=401.0                  & 2.27            & 28.64              & 69.09    & 220           \\ \hline
4                    & admission\_type\_id=1; discharge\_disposition\_id=3                          & 14.76           & 34.35              & 50.89    & 7813          \\ \hline
5                    & admission\_type\_id=1; discharge\_disposition\_id=3; admission\_source\_id=7 & 15.09           & 35.41              & 49.50    & 6645          \\ \hline
6                    & discharge\_disposition\_id=3; admission\_source\_id=7                        & 15.19           & 36.97              & 47.84    & 8290          \\ \hline
\end{tabular}
\label{tab:par}
\end{table*}

\subsection{ Cost Analysis}
The cost analysis has been done as explained in the section 3.E. The article \cite{30} specifies that  cost of readmission of Diabetes mellitus and its complications to be \$251 million for 23,700 total readmissions. Hence cost per readmission \(\alpha\) approximately equals to \$10,591 (\begin{math}\$250,000,000/23,300\end{math}). In our research, special diagnosis is considered to be One Extra Day of admission during which the doctor can conduct another diagnosis which might result in the discovery of new complications in the patient. Hence the cost of special diagnosis is considered to be cost per one day of admission in the hospital. From our dataset we find that the average time in hospital across diabetic patient encounters is 4.396 days. Hence cost for one day admission is considered to be \$2,409 (\begin{math}\$10591/4.396\end{math}). Hence the value of \(\alpha\) and \(\beta\) are \$10,591 and \$2,409 respectively.

Hence the saved cost matrix would be: 
\begin{equation}
Saved Cost Matrix (S) = 
\begin{bmatrix}
\alpha - \beta    &     0     \\
-\beta      &     0     \\
\end{bmatrix}
=
\begin{bmatrix}
8182  &     0     \\
-2409 &     0     \\
\end{bmatrix}
\end{equation}

Model with different machine learning algorithms were tuned to maximize the cost saved by selecting appropriate threshold. The cost saved by different models are given the Table ~\ref{tab:costsavingcomp}.

\begin{table}[h]
\centering
\caption{Comparison of cost saved by different machine learning models.}
\begin{tabular}{|l|l|l|}
\hline
\textbf{Algorithm} & \textbf{\begin{tabular}[c]{@{}l@{}}Cost Saved for \\ Test Set \\ (Million USD)\end{tabular}} & \textbf{\begin{tabular}[c]{@{}l@{}}Cost Saved for \\Total set\\   (Million USD)\end{tabular}} \\ \hline
Naive Bayes        & 58.726                                                                                       & 249.783563                                                                                     \\ \hline
Bayes Network      & 58.808                                                                                       & 250.1323396                                                                                    \\ \hline
Random Forest      & 59.425                                                                                       & 252.7566705                                                                                    \\ \hline
Adaboost-Trees     & 58.298                                                                                       & 247.9631195                                                                                    \\ \hline
Neural Network     & 58.963                                                                                       & 250.7916123                                                                                    \\ \hline
\end{tabular}
\label{tab:costsavingcomp}
\end{table}

We observe that the Random Forest model would save maximum cost of \$59.425 million for 23,053 diabetic patient encounter instances of test set. We get \$252.76 million by extending this to total number of 98,053 diabetic patient encounter instances. 

\section{Conclusion and Future Work}
In this work we presented a scheme to identify high-risk patients and evaluated different machine-learning algorithms. In contrast to previous work we considered both short-term and long-term readmissions and focussed on specific disease like diabetes. We found that random forests were optimal for this task. The dataset of readmissions is often skewed and consequently, the performance of identifying high-risk patients is modest. We conducted some preliminary experiments to modify the loss function of the classifiers to account for the skewed nature of the dataset. Slight improvement in the performance was observed comapared to previous works but not significant enough. Larger datasets containing medical records of readmitted patients are likely to be helpful for future research. Moreover, this work uncovers the features that are critical in identifying high risk of readmission and compares them in short-term and long-term analysis. In addition, statistically significant associations taht exist among these features are presented. TO complement the prediciton and feature analysis model, a cost sensitive model has been proposed.     

From the cost analysis, we observe that a cost of \$252.76 million can be saved for 98,053 instances of diabetic patient encounters. Saving such huge amount cost is essential for healthcare system. The model never suggests healthcare personnel to give less attention to those patients predicted not to be readmitted, but prompts extra attention to those who were predicted to get readmitted. In this sense, the designed model is conservative in nature and is safe to use in healthcare institutions as it enhances preventive strategy along with saving cost associated with readmission. Moreover, those patients who were predicted to be readmitted would receive special diagnosis at an earlier stage which might save lot of lives. Hence we believe that the model could be incorporated in healthcare institutions to witness its effectiveness.

Our research considers the hypothesis that the special diagnosis would prevent the actual readmission. We have considered cost of special diagnosis to be cost per one day of admission where doctor can run another diagnosis to discover other possible complications. Though these hypotheses seem legitimate they need to be tested by incorporating the research model in real healthcare systems. Extensive study needs to be done on the feature importance analysis which would help the healthcare institutions to prioritize their healthcare data documentation system. The cost saved by correctly predicting the patients who do not get readmitted might save some cost as it helps in optimal resource planning, but this cost needs to statistically determined, and the model should also be tuned by considering this cost into account. 

Our research targets diabetic patients only. Such analysis needs to be carried for other top health conditions like Heart disease, Schizophrenia etc. In the future studies scheduled and unscheduled readmissions needs to be considered \cite{24}. Several critical features in the medical records, like age were found missing. Hence a superior data collection drive needs to done for future research in this regard. Some features that needs to collected would be age, date of admission (to find the season of the year), number of patients with same disease at the instance of admission (to co-relate to epidemics), family history (to find hereditary information). The conversation between doctor and patient could also be collected which might help to extract essential features corresponding to patients will power and attitude by text-mining techniques which in turn might improve the intelligent models to identify patients with high risk of readmission. 

%
\section*{Acknowledgment}
We extend our immense gratitude to Prof. Bhiksha Raj, Prof. Rita Singh (Language Technology Institute, School of Computer Science, Carnegie Mellon University) and Pulkit Agrawal (Dept. of Computer Science, University of California, Berkley) for their keen guidance enabling the project.
We would like to thank M. Lichman\cite{3} for making the dataset available in the UCI machine learning library. 

\appendices
\section{Significant Association Rules}
This section contains Table ~\ref{tab:appendixpar} presenting a selected list of association rules.  They can be interpreted as explained in section 4.B.2. These rules can be used as suggestions by a doctor. Tables ~\ref{tab:appendixicd9} and ~\ref{tab:appendixmapping}  describe the ICD9 code mappings for the ICD9 codes and the Id mappings for the Id's used in Table ~\ref{tab:par} respectively. Complete mappings can be found in \cite{3}.

\begin{table*}[h]
\centering
\caption{Prominent Association Rules helpful for identifying patient encounters which are Not \textless30 day readmissions}
\begin{tabular}{|l|l|l|l|l|l|}
\hline
\multirow{3}{*}{No.} & Readmission Status                                                                                                       & \textless30     & \textgreater30     & NO       & TOTAL         \\ \cline{2-6} 
                     & Total Number of Instances                                                                                                & 11357           & 35545              & 54863    & 101765        \\ \cline{2-6} 
                     & \begin{tabular}[c]{@{}l@{}}Association Rules\\ (Consequents)\end{tabular}                                                & \multicolumn{3}{l|}{Class-Wise Matches (in \%)} & Total Matches \\ \hline
1                    & diag\_1=682.0; diag\_2=250.0                                                 & 1.88            & 26.29              & 71.83    & 213           \\ \hline
2                    & diag\_1=786.0; diag\_2=250.0; diag\_3=401.0                                  & 2.16            & 22.51              & 75.32    & 231           \\ \hline
3                    & race=Caucasian; gender=Female; diag\_1=786.0; diag\_3=401.0                  & 2.27            & 28.64              & 69.09    & 220           \\ \hline
4                    & race=Caucasian; diag\_1=386.0                                                & 2.48            & 42.15              & 55.37    & 121           \\ \hline
5                    & race=AfricanAmerican; gender=Female; diag\_1=786.0; diag\_3=250.0            & 2.68            & 34.90              & 62.42    & 149           \\ \hline
6                    & diag\_2=493.0; diag\_3=250.0                                                 & 2.70            & 36.49              & 60.81    & 148           \\ \hline
7                    & diag\_2=411.0; diag\_3=V45                                                   & 2.83            & 48.11              & 49.06    & 106           \\ \hline
8                    & diag\_1=414.0; diag\_2=411.0; diag\_3=V45                                    & 2.97            & 46.53              & 50.50    & 101           \\ \hline
9                    & race=Caucasian; gender=Female; diag\_2=414.0; diag\_3=V45                    & 3.60            & 44.14              & 52.25    & 111           \\ \hline
10                   & race=Caucasian; diag\_1=414.0; diag\_3=414.0                                 & 3.61            & 39.69              & 56.70    & 194           \\ \hline
11                   & gender=Female; diag\_2=414.0; diag\_3=V45                                    & 3.68            & 42.65              & 53.68    & 136           \\ \hline
12                   & race=Caucasian; gender=Male; diag\_3=413.0                                   & 3.70            & 40.74              & 55.56    & 135           \\ \hline
13                   & race=Caucasian; gender=Female; diag\_1=486.0; diag\_3=428.0                  & 4.26            & 50.35              & 45.39    & 141           \\ \hline
14                   & gender=Female; diag\_1=486.0; diag\_3=428.0                                  & 4.65            & 51.16              & 44.19    & 172           \\ \hline
15                   & discharge\_disposition\_id=1; admission\_source\_id=1                        & 8.76            & 32.46              & 58.79    & 18812         \\ \hline
16                   & admission\_type\_id=1; discharge\_disposition\_id=1                          & 9.60            & 37.31              & 53.09    & 31695         \\ \hline
17                   & admission\_type\_id=2; discharge\_disposition\_id=1                          & 9.69            & 34.68              & 55.63    & 11700         \\ \hline
18                   & admission\_type\_id=1; discharge\_disposition\_id=1; admission\_source\_id=7 & 9.78            & 38.44              & 51.79    & 28726         \\ \hline
19                   & discharge\_disposition\_id=1; admission\_source\_id=7                        & 9.93            & 39.13              & 50.93    & 33981         \\ \hline
20                   & A1Cresult=None; insulin=No                                                   & 10.23           & 34.19              & 55.58    & 39978         \\ \hline
21                   & admission\_type\_id=3; admission\_source\_id=1                               & 10.24           & 30.39              & 59.37    & 16187         \\ \hline
22                   & insulin=No; change=No                                                        & 10.25           & 33.46              & 56.28    & 37338         \\ \hline
23                   & A1Cresult=None; insulin=No; change=No                                        & 10.35           & 33.67              & 55.98    & 32786         \\ \hline
24                   & A1Cresult=None; change=No                                                    & 10.87           & 34.04              & 55.08    & 45674         \\ \hline
25                   & race=Caucasian; gender=Male                                                  & 11.07           & 34.70              & 54.22    & 36410         \\ \hline
26                   & race=AfricanAmerican; gender=Female                                          & 11.08           & 34.98              & 53.94    & 11728         \\ \hline
27                   & admission\_type\_id=2; admission\_source\_id=1                               & 11.16           & 33.14              & 55.70    & 9224          \\ \hline
28                   & diag\_2=H; A1Cresult=None                                                    & 11.25           & 36.93              & 51.81    & 25870         \\ \hline
29                   & race=Caucasian; gender=Female                                                & 11.49           & 36.50              & 52.01    & 39689         \\ \hline
30                   & diag\_1=H; A1Cresult=None                                                    & 11.62           & 36.15              & 52.23    & 24397         \\ \hline
31                   & A1Cresult=None; insulin=Steady                                               & 11.69           & 34.97              & 53.34    & 24269         \\ \hline
32                   & A1Cresult=None; change=Ch                                                    & 12.35           & 37.22              & 50.43    & 36186         \\ \hline
33                   & discharge\_disposition\_id=6; admission\_source\_id=7                        & 14.42           & 44.15              & 41.43    & 6840          \\ \hline
34                   & admission\_type\_id=1; discharge\_disposition\_id=3                          & 14.76           & 34.35              & 50.89    & 7813          \\ \hline
35                   & admission\_type\_id=1; discharge\_disposition\_id=3; admission\_source\_id=7 & 15.09           & 35.41              & 49.50    & 6645          \\ \hline
36                   & discharge\_disposition\_id=3; admission\_source\_id=7                        & 15.19           & 36.97              & 47.84    & 8290          \\ \hline
\end{tabular}
\label{tab:appendixpar}
\end{table*}

\begin{table*}[h]
\centering
\caption{Prominent ICD9 Code Mappings}
\begin{tabular}{|l|l|}
\hline
Diagnoses (ICD9 Codes) & Diseases                                          \\ \hline
682.0                  & Cellulitis/abscess, face                          \\ \hline
250.0                  & Diabetes mellitus without mention of complication \\ \hline
786.0                  & Dyspnea and respiratory abnormalities             \\ \hline
401.0                  & Hypertension, malignant                           \\ \hline
\end{tabular}
\label{tab:appendixicd9}
\end{table*}

\begin{table*}[h]
\centering
\caption{Prominent Admission Source, Type and Discharge Disposition Id Mappings}
\begin{tabular}{|l|l|}
\hline
Identifiers                   & Mappings                                   \\ \hline
admission\_type\_id=1         & Emergency                                  \\ \hline
admission\_source\_id=7       & Emergency Room                             \\ \hline
discharge\_disposition\_id=3; & Discharged or Transferred to Skilled Nursing Facility (SNF) \\ \hline
\end{tabular}
\label{tab:appendixmapping}
\end{table*}

\ifCLASSOPTIONcaptionsoff
  \newpage
\fi

\bibliographystyle{abbrv}
\bibliography{main} 

\end{document}